\documentclass[conference]{IEEEtran}
\IEEEoverridecommandlockouts
\usepackage{cite}
\usepackage{amsmath,amssymb,amsfonts}
\usepackage{tabulary,graphicx,times,caption,fancyhdr,amsfonts,amssymb,amsbsy,latexsym,amsmath}
\usepackage[utf8]{inputenc}
\usepackage{url,multirow,morefloats,floatflt,cancel,tfrupee,textcomp,colortbl,xcolor,pifont}
\usepackage[ruled]{algorithm2e}
\usepackage{graphicx}
\usepackage{textcomp}
\usepackage{xcolor}
\usepackage{multicol}
\usepackage{url}
\usepackage{cite}
\usepackage{hyperref} 
\usepackage{booktabs}
\newcommand{\Tau}{\mathcal{T}}

\newcommand{\N}{\mathcal{N}}

\newcommand{\E}{\operatorname{\mathbb{E}}}

\def\BibTeX{{\rm B\kern-.05em{\sc i\kern-.025em b}\kern-.08em
    T\kern-.1667em\lower.7ex\hbox{E}\kern-.125emX}}
\begin{document}

\title{A Decision-Making GPT Model Augmented with Entropy Regularization for Autonomous Vehicles\\
\thanks{This work was supported in part by the National Key R\&D Program of China (2023YFB4301900), the Shanghai Scientific Innovation Foundation (No.23DZ1203400), the National Natural Science Foundation of China (52302502), the State Key Laboratory of Intelligent Green Vehicle and Mobility under Project No. KFZ2408, the Young Elite Scientists Sponsorship Program by CAST (2022QNRC001), and the Fundamental Research Funds for the Central Universities.}
\thanks{Corresponding author: Peng Hang} 
}

\author{\IEEEauthorblockN{1\textsuperscript{st} Jiaqi Liu}
\IEEEauthorblockA{
\textit{Department of Traffic Engineering  \&} \\
\textit{Key Laboratory of} \\
\textit{Road and Traffic Engineering,} \\
\textit{Ministry of Education} \\
\textit{Tongji University}\\
Shanghai, China \\
liujiaqi13@tongji.edu.cn}

\and
\IEEEauthorblockN{2\textsuperscript{nd} Shiyu Fang}
\IEEEauthorblockA{
\textit{Department of Traffic Engineering  \&} \\
\textit{Key Laboratory of} \\
\textit{Road and Traffic Engineering,} \\
\textit{Ministry of Education} \\
\textit{Tongji University}\\
Shanghai, China \\
2111219@tongji.edu.cn}

\and
\IEEEauthorblockN{3\textsuperscript{nd} Xuekai Liu}
\IEEEauthorblockA{
\textit{Department of Traffic Engineering  \&} \\
\textit{Key Laboratory of} \\
\textit{Road and Traffic Engineering,} \\
\textit{Ministry of Education} \\
\textit{Tongji University}\\
Shanghai, China \\
2310795@tongji.edu.cn}

\and
\IEEEauthorblockN{4\textsuperscript{nd} Lulu Guo}
\IEEEauthorblockA{
\textit{Department of Control Science} \\
\textit{and Engineering} \\
\textit{Tongji University}\\
Shanghai, China \\
guoll21@tongji.edu.cn}

\and
\IEEEauthorblockN{5\textsuperscript{nd} Peng Hang}
\IEEEauthorblockA{
\textit{Department of Traffic Engineering  \&} \\
\textit{Key Laboratory of} \\
\textit{Road and Traffic Engineering,} \\
\textit{Ministry of Education} \\
\textit{Tongji University}\\
Shanghai, China \\
hangpeng@tongji.edu.cn}

\and
\IEEEauthorblockN{6\textsuperscript{rd} Jian Sun}
\IEEEauthorblockA{
\textit{Department of Traffic Engineering  \&} \\
\textit{Key Laboratory of} \\
\textit{Road and Traffic Engineering,} \\
\textit{Ministry of Education} \\
\textit{Tongji University}\\
Shanghai, China \\
sunjian@tongji.edu.cn}

}

\maketitle

\begin{abstract}
In the domain of autonomous vehicles (AVs), decision-making is a critical factor that significantly influences the efficacy of autonomous navigation. As the field progresses, the enhancement of decision-making capabilities in complex environments has become a central area of research within data-driven methodologies. Despite notable advances, existing learning-based decision-making strategies in autonomous vehicles continue to reveal opportunities for further refinement, particularly in the articulation of policies and the assurance of safety. In this study, the decision-making challenges associated with autonomous vehicles are conceptualized through the framework of the Constrained Markov Decision Process (CMDP) and approached as a sequence modeling problem. Utilizing the Generative Pre-trained Transformer (GPT), we introduce a novel decision-making model tailored for AVs, which incorporates entropy regularization techniques to bolster exploration and enhance safety performance. Comprehensive experiments conducted across various scenarios affirm that our approach surpasses several established baseline methods, particularly in terms of safety and overall efficacy.
\end{abstract}


\section{Introduction}
Autonomous vehicles (AVs) are poised to fundamentally transform the transportation landscape, promising enhanced safety and efficiency for all road users~\cite{hang2022conflict}. At the heart of these systems, the decision-making module is crucial for the overall performance of AVs~\cite{liu2024enhancing}. Despite significant advances, optimizing the decision-making capabilities of AVs under complex traffic scenarios remains a formidable challenge, particularly in enhancing safety and generalization abilities.

The rapid advancement of artificial intelligence (AI) technologies offers promising solutions to these challenges. Data-driven methods, notably reinforcement learning (RL), have emerged as potent tools in this regard~\cite{kiran2021deep, liu2023towards}. RL, a well-established and effective approach for sequential decision-making, trains agents to maximize cumulative rewards through interaction with their environments, demonstrating a robust capacity to capture dynamic interrelationships and identify optimal driving strategies~\cite{liu2023teaching}. Nevertheless, traditional RL methods—both online and offline—still grapple with significant issues, including safety, sampling efficiency, and generalization~\cite{kiran2021deep}.

In addition to traditional RL techniques, Transformer-based methods have shown exceptional performance across various AI tasks, such as Natural Language Processing (NLP) and Computer Vision~\cite{casola2022pre}. These methods are increasingly being recognized as powerful alternatives for modeling complex decision-making problems~\cite{chen2021decision,zheng2022online,liu2023mtd,liu2023constrained}. Liu et al.~\cite{liu2023mtd} have proposed the Decision-Making GPT model, utilizing GPT to address multi-task decision-making challenges. Their findings suggest that the Decision-Making GPT model outperforms traditional RL-trained specialist models, significantly enhancing the generalization capabilities of data-driven decision-making approaches. Despite the demonstrated efficacy of the GPT model in decision-making scenarios, there remains a pressing need to further enhance the safety and overall performance of Transformer-based decision-making methods.

\begin{figure}
    \centering
    \includegraphics[width=1\linewidth]{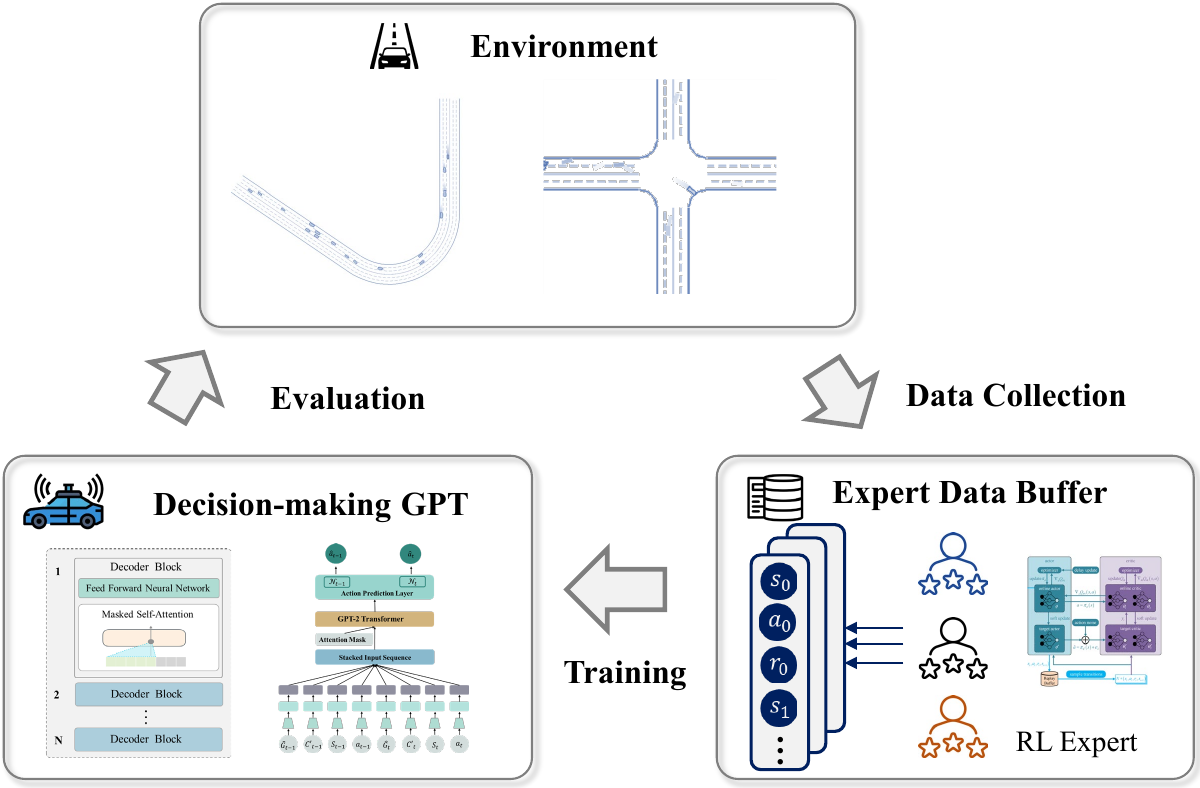}
    \caption{The overall procedure of our work.}
    \label{fig:overall_procedure}
\end{figure}

Building upon the contextual framework outlined, our research extends previous efforts by first conceptualizing the decision-making challenges in autonomous vehicles as a Constrained Markov Decision Process (CMDP) and simultaneously addressing these as a sequence modeling issue. We harness the capabilities of the Generative Pre-trained Transformer (GPT) model, specifically leveraging insights from the architecture of GPT-2~\cite{radford2019language}, to learn from human expert driving data and to interact effectively with the environment. We propose an improved GPT model, incorporating sequence-level entropy regularizers aimed at improving both safety and sampling efficiency.
This improved GPT model, designed for end-to-end driving tasks, learns from expert decision-making data and makes auto-regressive driving decisions akin to those in natural language processing challenges. Moreover, it employs a stochastic policy representation with entropy regularization as its optimization target. This approach mitigates the compounding errors typically observed in offline settings and permits the policy to explore a broader range of actions, thus significantly boosting the model's overall performance. Additionally, an offline expert data collection module has been developed to train multiple reinforcement learning agents, accumulating valuable data across varied scenarios. The comprehensive workflow of this process is depicted in Figure \ref{fig:overall_procedure}.

Furthermore, we trained and evaluated GPT models with varying parameter sizes across different traffic scenarios. The results demonstrate that our model achieves superior safety and overall performance compared to both the baseline methods and the original decision-making GPT model.

The contributions of our work are summarized as follows:
\begin{itemize}
    \item We abstract and model the decision-making problem for autonomous vehicles as a sequence modeling and prediction task, utilizing the GPT model.
    \item We introduce an improved decision-making GPT model based on the GPT-2 architecture, employing a Shannon entropy regularizer to enhance safety performance.
    \item Extensive experiments confirm that our enhanced GPT model surpasses other baseline methods in terms of decision-making quality and safety performance.
\end{itemize}

\section{Preliminaries}
This section introduces foundational concepts relevant to the constrained Markov decision process and safe reinforcement learning, followed by the formalization of the decision-making problem at hand.

\subsection{Constrained Markov Decision Process (CMDP) and Safe RL}
Safe reinforcement learning (safe RL) represents a subset of reinforcement learning (RL) methodologies designed to enhance the safety of RL algorithms, typically framed within the Constrained Markov Decision Process (CMDP) paradigm~\cite{altman2021constrained,liu2024ddm}. A CMDP, defined as a finite horizon $\mathcal{M}$, consists of the tuple $(\mathcal{S}, \mathcal{A}, \mathcal{P}, r, c, \mu_0)$. Here, $\mathcal{S}$ denotes the state space, $\mathcal{A}$ the action space, $\mathcal{P}: \mathcal{S} \times \mathcal{A} \times \mathcal{S} \rightarrow [0, 1]$ the transition probability function, $r: \mathcal{S} \times \mathcal{A} \times \mathcal{S} \rightarrow \mathbb{R}$ the reward function, and $\mu_0: \mathcal{S} \rightarrow [0,1]$ the initial state distribution. The CMDP framework extends the traditional Markov Decision Process (MDP) by incorporating an additional cost function $c: \mathcal{S} \times \mathcal{A} \times \mathcal{S} \rightarrow [0, C_{max}]$, which assigns a cost for constraint violations, with $C_{max}$ representing the maximum possible cost. Although this framework can be applied to scenarios with multiple constraints or partial observability, for simplicity, this study focuses on CMDPs with a single, explicit constraint.

The objective of safe RL within this CMDP framework is defined by both a CMDP and a constraint threshold $\kappa \rightarrow [0, +\infty)$. A policy $\pi: \mathcal{S} \times \mathcal{A} \rightarrow [0,1]$ governs the action selection, and a trajectory $\tau = \{s_1, a_1, r_1, c_1, ..., s_T, a_T, r_T, c_T\}$, where $T = |\tau|$ is the maximum episode length. The cumulative reward $R(\tau) = \sum_{t=1}^{T} r_t$ and the total cost $C(\tau)=\sum_{t=1}^{T} c_t$ of a trajectory $\tau$ are used to evaluate performance. The principal aim of safe RL is to optimize the policy that maximizes the expected reward, constrained by the requirement that the expected cost does not exceed the threshold $\kappa$:
\begin{equation}
\vspace{-2mm}
\max_{\pi} \mathbb{E}{\tau \sim \pi} \big[R(\tau) \big], \quad \text{s.t.} \quad \mathbb{E}{\tau \sim \pi} \big[C(\tau) \big] \leq \kappa.
\label{eq:safe-rl}
\end{equation}
In an offline setting, where the agent cannot collect additional data through interaction but must rely on pre-collected trajectories from potentially arbitrary and unknown policies, this formulation poses unique challenges to achieving the constrained optimization objectives.

\subsection{Problem Formulation}
This subsection elucidates the formulation of our problem by precisely defining the state and action spaces used within our model.
\subsubsection{State Space}
The state input $S$ of our AV model is delineated into four primary components. The initial component captures the ego vehicle's own state, characterized by its position \( [x_{\text{ego}}, y_{\text{ego}}] \), velocity \( [v_{x_{\text{ego}}}, v_{y_{\text{ego}}}] \), steering angle $[heading]$, and the distance from the road boundary $[dis_{bound}]$. The second component involves navigation information, wherein a route from the origin to the destination is computed, and checkpoints are established at defined intervals. This setup provides navigation data, including the relative distance and direction to the subsequent checkpoint. The third component is a 240-dimensional vector that effectively models the vehicle's surrounding environment, similar to the data obtained from LiDAR point clouds. This data is gathered through a LiDAR sensor that scans a 360-degree horizontal field of view with 240 lasers, a maximum detection radius of 50 meters, and a horizontal resolution of 1.5 degrees. The final component integrates the state of surrounding vehicles, including their positions and velocities, acquired through V2X communication technologies.

\subsubsection{Action Space}
The action space of our model, represented by \( \mathbf{A} = [a_1, a_2]^T \in (0,1) \), is confined to the interval (0,1) and is employed to manage the lateral and longitudinal motions of the vehicle. These normalized actions are translated into specific low-level continuous control commands as follows: \begin{equation}
    \label{action}
    \begin{aligned}
        &u_s = S_{\text{max}} \cdot a_1\\
        &u_a = F_{\text{max}} \cdot \max\{0, a_2\}\\ 
        &u_b = -B_{\text{max}} \cdot \min\{0, a_2\}
    \end{aligned}
\end{equation}
\noindent Here, \( S_{\text{max}} \) denotes the maximum steering angle, \( F_{\text{max}} \) specifies the maximum engine force, and \( B_{\text{max}} \) represents the maximum braking force.

\section{Methodology}
This section delineates the architecture of our decision-making model, focusing on the incorporation of entropy regularization to enhance the performance of the GPT model.

\subsection{Decision-making GPT Model}
In alignment with prior work \cite{liu2023mtd}, the training process of the decision-making GPT is conceptualized as a sequence modeling challenge and is conducted in an autoregressive manner.

\subsubsection{Input Representation with Target Cost Threshold}
Consider a trajectory $\tau \ (\tau \in \mathcal{D})$ representing a sequence of actions taken by an AV, sampled from the expert offline dataset, with $|\tau|$ denoting its length. The Return-to-Go (RTG) for the trajectory at timestep $t$ is defined as: $g_t = \sum^T_{t' = t} r_{t'}$, which aggregates the future rewards of the AV from timestep $t$ onward. Distinct from the approach in \cite{liu2023mtd}, we introduce an additional element representing the target cost threshold $c^\prime$, enhancing the input sequence. Here, $c^\prime_t=\sum_{t'=t}^T c_{t'}$ reflects the accumulated cost from timestep $t$.
Define $\mathbf{s}=(s_1,...,s_{|\tau|})$, $\mathbf{a} = (a_1,...,a_{|\tau|})$, $\mathbf{g} = (g_1,...,g_{|\tau|})$, and $\mathbf{c^\prime} = (c_1,...,c_{|\tau|})$ as the sequences representing state, action, RTG, and cost for $\tau$, respectively. The composite representation of a trajectory that is input to the GPT model is thus formatted as:
\begin{equation}
\tau^{\prime} = \Big( s_1, a_1, g_1, c_1, s_2, a_2, g_2, c_2, ..., s_T, a_T, g_T, c_T \Big)
\end{equation}
The initial RTG, $g_1$, corresponds to the total expected return of the trajectory. 

\subsubsection{Architecture}

\begin{figure}
    \centering
    \includegraphics[width=0.45\textwidth]{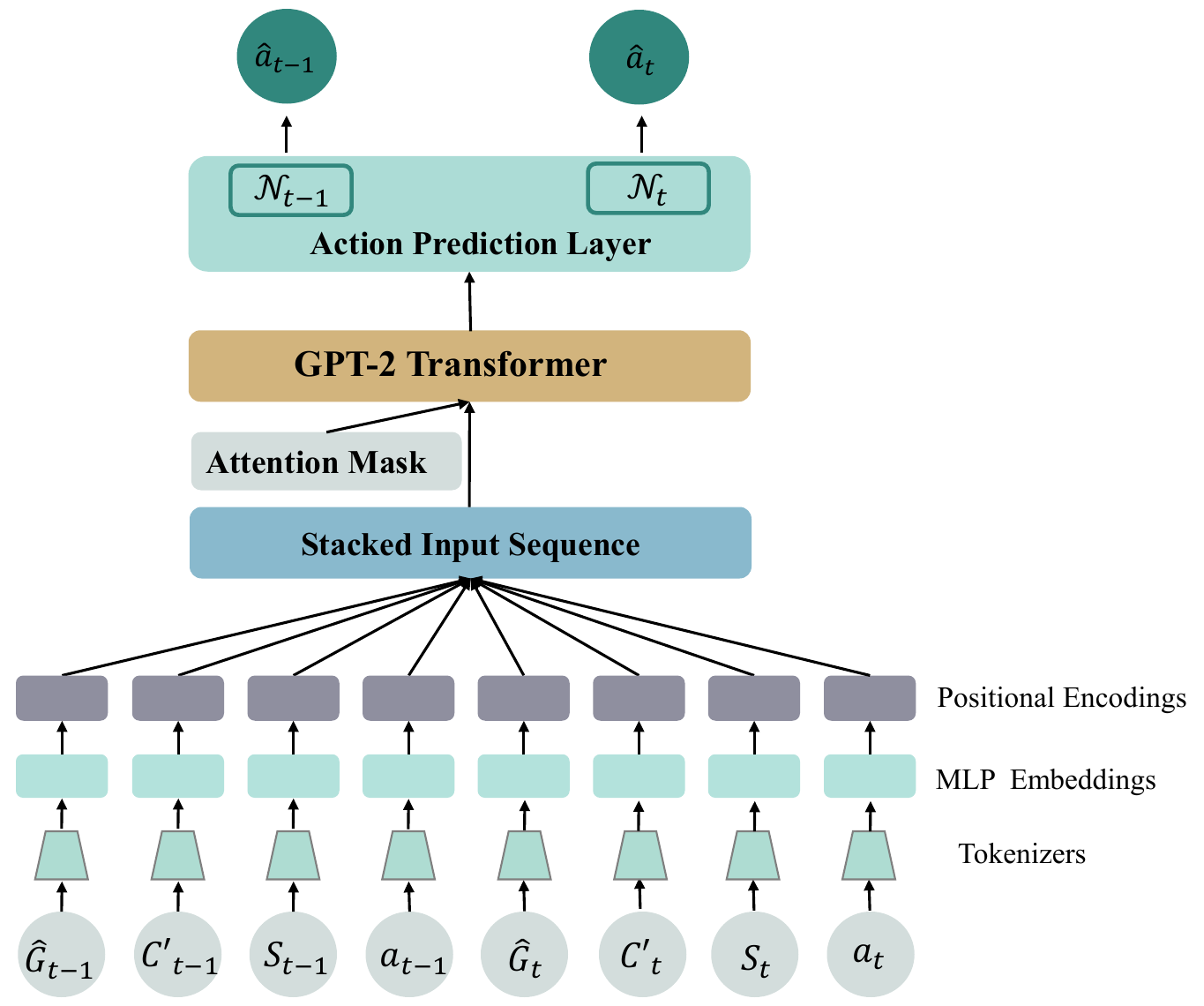}
    \caption{The workflow of our decision-making GPT model.}
    \label{fig:Model_GPT}
\end{figure}

As shown in Figure \ref{fig:Model_GPT}, the decision-making GPT model utilizes an approach similar to NLP techniques for modeling decision-making tasks. 

Initially, expert trajectories $\tau^{\prime}$ are randomly sampled from the data pool $\mathcal{D}$, and these trajectories are then transformed into tokens suitable for the model's processing.
Denote $\mathbf{x}_t =\{ \mathbf{s}_{-K:t}, \mathbf{a}_{-K:t-1},\mathbf{g}_{-K:t}, \mathbf{c^\prime}_{-K:t} \}$ as the input tokens at timestep $t$, where $\mathbf{s}_{-K,t}$ is shorthand for the sequence of $K$ past states $\mathbf{s}_{\max (1,t-K+1):t}$, similarly for $\mathbf{a}_{-K:t-1}$,$\mathbf{g}_{-K,t}$ and $\mathbf{c^\prime}_{-K,t}$. $K$ is a hyperparameter and is also referred to as the $context \ length$ for the GPT.

Subsequently, a Multilayer Perceptron (MLP) is employed to map these tokens into a continuous vector space. Positional encodings (PE), as proposed in \cite{vaswani2017attention}, are added to these embeddings to preserve the sequential order of the input tokens:
\begin{equation}
    \mathbf{x}^{\prime}_t = \text{MLP}(\mathbf{x}_x)
\end{equation}
\begin{equation}
    \mathbf{e}_t = \mathbf{x}^{\prime}_t + \text{Positional\_Encoding}(\mathbf{x}_t)
\end{equation}

These embeddings are subsequently fed into the Transformer layers, where they undergo transformations to produce hidden states $\mathbf{h}_t$:
\begin{equation}
    \mathbf{h_t} = \text{TransformerLayer}(\mathbf{e}_t)
\end{equation}

Finally, a linear prediction layer is used to generate predicted action $a^{\prime}_{t+1}$ based on the hidden states:
\begin{equation}
    a^{\prime}_{t+1} = \text{Linear}(\mathbf{h}_t)
\end{equation}

\subsection{Sequence Modeling with Entropy Regularization}

This subsection delineates the learning objective with entropy regularization for our decision-making GPT model, emphasizing the development of a probabilistic, stochastic policy that optimizes the likelihood of actions from the expert dataset in a continuous action space.

We adopt a generalized probabilistic learning objective and extend it to incorporate exploration strategies into our decision-making framework. In alignment with standard practices in the field \cite{zheng2022online,haarnoja2018soft}, we model the action distributions conditioned on states and RTGs using a multivariate Gaussian distribution with a diagonal covariance matrix. Formally, the policy of the decision-making GPT, denoted by $\pi$ with parameters $\theta$, is defined as follows:
\begin{equation}
\begin{aligned}
& && \pi_\theta(a_t | \mathbf{o}_t )\\
& = && \N(\mu_\theta(\mathbf{o}_t ), \Sigma_\theta(\mathbf{o}_t )), \; \forall t,
\end{aligned}
\end{equation}
where the covariance matrix $\Sigma_\theta$ is assumed to be diagonal and $\mathbf{o}_t = \{ s_{-K:t},g_{-K:t}, c^\prime_{-K:t} \} $.
Given a stochastic policy, we aim to maximize the log-likelihood of observed trajectories, equivalently minimizing the negative log-likelihood (NLL) loss as follows:
\begin{equation}
    \begin{aligned}
           & J(\theta) = \tfrac{1}{K} \E_{ (\mathbf{a}, \mathbf{o}) \sim \mathcal{D} } [ - \log \pi_\theta(\mathbf{a} | \mathbf{o}) ] \\
     = &\,  \tfrac{1}{K} \E_{ (\mathbf{a}, \mathbf{o}) \sim \mathcal{D} } [ \textstyle -\sum_{k=1}^K \displaystyle \log \pi_\theta(a_k | \mathbf{o}_k)].
    \end{aligned}
    \label{eq:sdt_nll}
\end{equation}

To enhance exploration, we quantify it via the entropy of the policy, defined as:
\begin{equation}
\begin{aligned}
    & H_{\theta}[\mathbf{a} | \mathbf{o}]  = \tfrac{1}{K} \E_{(\mathbf{o}) \sim \mathcal{D}} \big[  H[\pi_\theta(\mathbf{a} | \mathbf{o} )]  \big] \\
    =  &  \tfrac{1}{K} \E_{(\mathbf{o}) \sim \Tau}\big[ \textstyle \sum_{k=1}^K H[\pi_\theta(a_k | \mathbf{o}_k) ]\big],
    \end{aligned}
    \label{eq:sdt_entropy}
\end{equation}
where $H[\pi_\theta(\mathbf{a})]$ denotes the Shannon entropy of the distribution $\pi_\theta(\mathbf{a})$. 

Aligning with prevalent max-entropy reinforcement learning algorithms~\cite{levine2018reinforcement}, we impose a lower bound on policy entropy to foster exploration. 
The optimization problem is thus formulated as:
\begin{equation}
    \min_\theta J(\theta) \;\; \text{subject to} \;\; H_{\theta}[\mathbf{a}| \mathbf{o} ] \geq \beta,
    \label{eq:sdt_main}
\end{equation}
where  $\beta$ is a prefixed hyperparameter.
Following \cite{haarnoja2018soft}, we address the dual problem of Equation \eqref{eq:sdt_main} to avoid direct handling of the inequality constraint, defining the Lagrangian as $L(\theta, \lambda) = J(\theta) + \lambda (\beta - H_{\theta} [\mathbf{a}| \mathbf{o}])$. The optimization alternates between minimizing $\theta$ and maximizing $\lambda$, structured as:
\begin{equation}
    \min_{\theta}\, J(\theta) - \lambda H_{\theta} [\mathbf{a}| \mathbf{o}],
    \label{eq:sdt_opt_theta}
\end{equation}
% and optimizing $\lambda$ with fixed $\theta$ boils down to solving
\begin{equation}
    \min_{\lambda \geq 0} \, \lambda (H_{\theta}[\mathbf{a}| \mathbf{o}] - \beta).
    \label{eq:sdt_opt_lambda}
\end{equation}

Consequently, the final loss function during training combines both NLL and entropy losses:
\begin{equation}
\begin{aligned}
    \ell_{\text{gpt}} = - \underbrace{\sum_{\mathbf{a}, \mathbf{o} \in \mathcal{D}}\log \pi_\theta (\mathbf{a} | \mathbf{o})}_{\ell_{\text{nll}}}
    - \underbrace{\lambda \sum_{\mathbf{o} \in \mathcal{D}} H[\pi_\theta(\cdot | \mathbf{o})]}_{\ell_{\text{ent}}}
\end{aligned}
\end{equation}

The whole training process of our model could be summarized in Algorithm \ref{algo:training}.

\begin{algorithm}
\SetAlFnt{\small} 
\SetKwInOut{Input}{Inputs}
\SetKwInOut{Output}{Outputs}
\caption{Decision-making GPT Model Training Procedure}
\label{algo:training}
\LinesNumbered
\SetAlgoLined
\Input{GPT model $\pi$, dataset $\mathcal{D}$, entropy weight $\lambda$, gradient steps $M$}
\Output{Trained GPT model $\pi_\theta$}
\vspace{0.2em}
\hrule
\vspace{0.2em}
Initialize GPT model with random weights $\theta$;\\
\For{$\text{Step} = 1$ to $M$}
{
    \textit{\bfseries  Sample offline data; }\\
    % $\mathcal{B} = \{\mathbf{a}_{i, t}, \mathbf{o}_{i, t} \}_{i=1}^B \sim \mathcal{T}, t \sim \text{SampleInt}(1, T)$;\\
    Sample a batch of trajectories $\tau$ with length $K$ from $\mathcal{D}$ with batch size $\mathcal{B}$ ;\\
    Compute RTG $g$ and cost $c^\prime$ for each trajectory $\tau$;\\
    Get input tokens $x$ : Tokenize$(\tau)$;\\
    \textit{\bfseries  Predict the next token; }\\
    Embed input tokens : $e = \text{MLP}(x) + PE_t$;\\
    Acquire hidden states $h$ by $\text{TransformerLayer}(e)$;\\
    Get predicted action $a$ by $\text{LinearLayer}(h)$;\\
    \textit{\bfseries  Compute the NLL loss and entropy loss; }\\
    $\ell_{\text{nll}} = -  \sum_{\mathbf{a}, \mathbf{o} \in \mathcal{D}}\log \pi_\theta (\mathbf{a} | \mathbf{o})$;\\  % \frac{1}{|\mathcal{B}|}
    $\ell_{\text{ent}} = - \sum_{\mathbf{o} \in \mathcal{D}} H[\pi_\theta(\cdot | \mathbf{o})]$;\\  %\frac{1}{|\mathcal{B}|}
    \textit{\bfseries Update the policy parameter; }\\
    $\ell_{\text{gpt}} =  \ell_{\text{nll}} + \lambda \ell_{\text{ent}}$;\\
    $\theta \leftarrow  \theta - \alpha \nabla_\theta\ell_{\text{gpt}}$;
}
\end{algorithm}

\section{Experiment and Evaluation}
In this section, the detailed information of the simulation environment and our models will be introduced. Sequently, the experiments results are analyzed.

\subsection{Experiments and Baselines}
\begin{figure}
    \centering
    \includegraphics[width=1\linewidth]{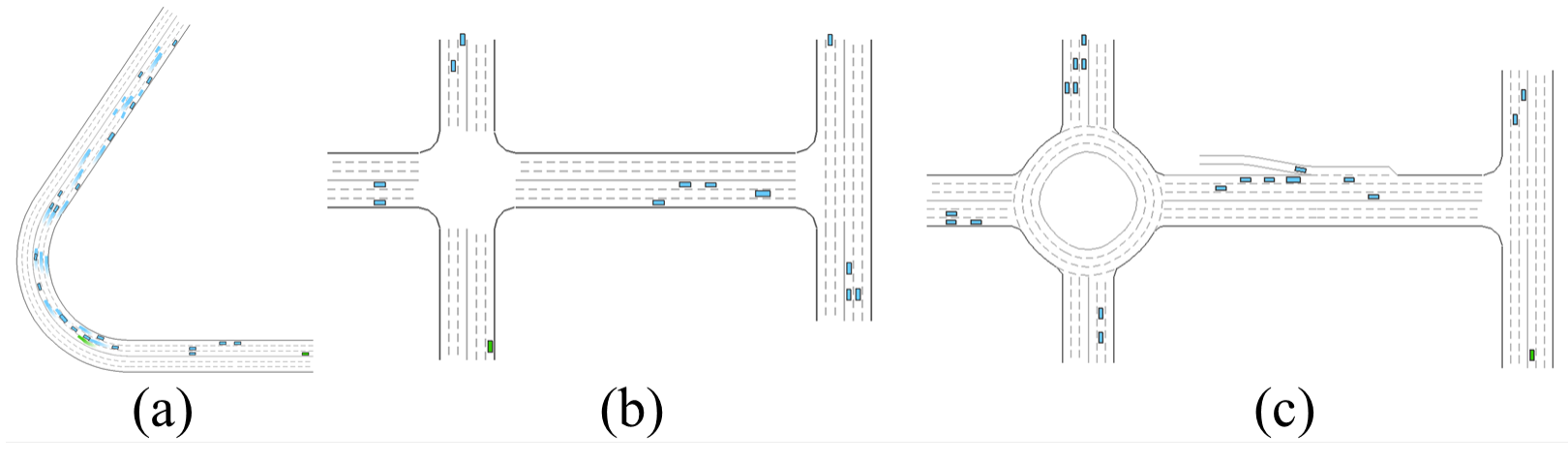}
    \caption{Three kinds of scenarios we used for our methods training and testing.}
    \label{fig:scenario_used}
\end{figure}

\textbf{Environments.}
Our experimental setup, encompassing data collection, training of the decision-making GPT model, and its evaluation, was conducted within the MetaDrive Simulator \cite{li2022metadrive}. This simulator, grounded in the OpenAI Gym Environment framework, facilitates the creation of diverse traffic scenarios. 
Our study encompasses three types of composite traffic scenarios: the first scenario integrates a straight road with a curved road, testing fundamental autonomous driving decision-making skills. The second scenario features two unsignalized intersections, one in a cross shape and another in a T shape, respectively. Lastly, the third scenario comprises a more intricate combination of a roundabout, a ramp-merge scenario, and a T-shaped intersection, posing greater challenges for autonomous vehicles in terms of interaction and decision-making. These scenarios are depicted in Figure \ref{fig:scenario_used}.

\begin{figure*}
    \centering
    \includegraphics[width=1\linewidth]{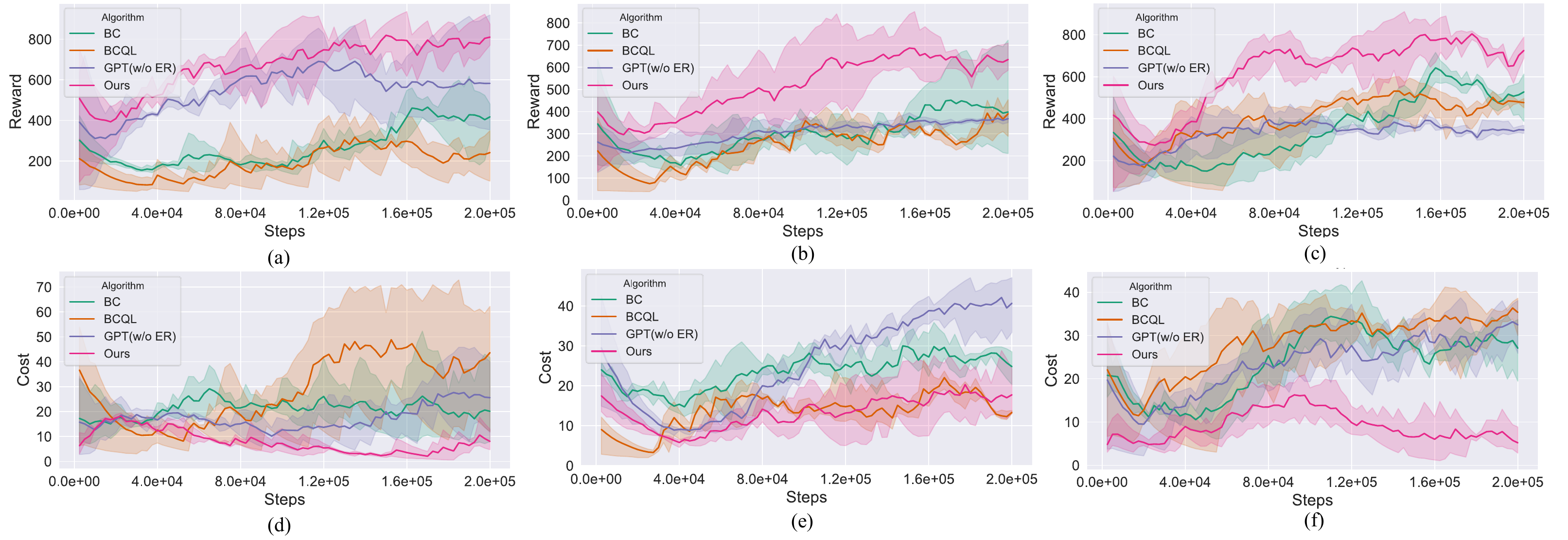}
    \caption{The reward and cost results of our method and baselines from different scenarios. Subfigures (a) and (d) are from scenario 1; subfigures (b) and (e) are from scenario 2; subfigures (c) and (f) are from scenario 3.}
    \label{fig:result_reward}
\end{figure*}

\textbf{Dataset Collection.}
The collection of a high-quality offline expert dataset is pivotal for the effective training of the decision-making GPT model. As depicted in Figure \ref{fig:overall_procedure}, we employ the CPPO algorithm \cite{stooke2020responsive}, an enhancement of the PPO-Lagrangian method, to accumulate experience data. This dataset includes trajectories and rewards from various environments, which are subsequently stored in a data buffer to form the offline dataset. For each scenario, differentiated by complexity and traffic density, approximately 10,000 trajectories are collected using the expert agent.

\textbf{Baselines.}
To establish a robust comparative framework, we selected several competitive baseline algorithms that are prevalent in the realms of safe RL and autonomous driving decision control. These include the classical Behavior Cloning algorithm (BC), the GPT model without entropy regularization \cite{chen2021decision}, and Batch-Constrained deep Q-learning (BCQL) \cite{fujimoto2019off}. These baselines provide a comprehensive reference point to evaluate the enhancements brought forth by our improved GPT model in handling complex driving decisions.

\subsection{Implementation Details}
For each model implemented in this study, training was conducted over $20,000$ timesteps with a batch size of 512. The specific parameters utilized for the decision-making GPT are detailed in Table \ref{tab:training_hyperparameter}.
The policy network of BC and BCQL are built based on a 3-layer MLP with 256 hidden units.
The reward and cost function of our expert data-collection model is defined as:
\begin{equation}
\label{eq:reward_function}
    R = \omega_1 r_{dis} + \omega_2 r_{v} + \omega_3 r_s
\end{equation}

\begin{equation}
\label{eq:cost_function}
    C = \omega'_{1} c_{1} + \omega'_{2} c_{2} + \omega'_{3} c_{3}
\end{equation}
The components include $r_{dis}$ for the reward based on distance covered, $r_{v}$ for the speed reward, and $r_{s}$ for the terminal reward. $c_{1}$, $c_{2}$ and $c_{3}$ are the penalty for the condition: out of road, crashing with other vehicles, crashing with other objects, respectively.
We set the following values for the reward function: $r_{dis}=1$, $r_{v}=0.1$, $r_{s}=10$, $r_{c1}=5.0$, $r_{c2}=5.0$ and $r_{c3}=5.0$. The coefficient of each reward term is set as 1.

All experiments were conducted in a computation platform with Intel Xeon Silver 4214R CPU and  NVIDIA GeForce RTX 3090 GPU $\times$ 2.

\begin{table}[!htbp]
    \centering
    \caption{The hyperparameter of the decision-making GPT model}
    \label{tab:training_hyperparameter}
    \begin{tabular}{c c c}
        \toprule
        Symbol & Definition & Value\\
        \midrule
        % $M$ & Training Epoch & 100 \\
        $L_r$ & Learning Rate & $10^{-4}$ \\
        $K$  & Training Context Length & 10 \\
        % $D$  & Dropout   & 0.1 \\
        $N_h$ & Number of Attention Heads & 8 \\
        % $N_a$ & Number of Action Head Layer & 1 \\
        $N_l$  & Number of Layers  &  3/6 \\
        $E_d$  & Embedding Dimension  &  128/ 512/ 1024 \\
        \bottomrule
    \end{tabular}
\end{table}

\subsection{Results Analysis}
Our methodology and various baselines were trained and evaluated across three distinct scenarios, utilizing both reward and safety cost as key performance indicators, as defined in Equations \ref{eq:reward_function} and \ref{eq:cost_function} respectively. The comparative training dynamics of all algorithms are depicted in Figure \ref{fig:result_reward}.

Figures \ref{fig:result_reward}(a), (b), and (c) present the training reward curves for each algorithm across the three scenarios. In Scenario 1, as shown in Figure \ref{fig:result_reward}(a), both our method and the GPT model without entropy regularization (w/o ER) demonstrate superior reward outcomes compared to the other two methods. However, in Scenarios 2 and 3, depicted in Figures \ref{fig:result_reward}(b) and (c) respectively, the GPT (w/o ER) fails to exhibit a competitive edge over the BC and BAQL algorithms, suggesting challenges in managing complex and highly interactive driving tasks. Nonetheless, our enhanced approach consistently delivers the highest rewards and optimal operational performance across all scenarios.

In terms of safety performance, Figures \ref{fig:result_reward}(d), (e), and (f) illustrate the safety costs incurred in each scenario, reflecting the incidence of safety violations. While the BCQL algorithm shows commendable safety performance in Scenario 2, it nevertheless incurs substantial safety violations in Scenarios 1 and 3. Similarly, both BC and GPT (w/o ER) exhibit deficiencies across varying scenarios. In contrast, our method consistently outperforms the baselines by achieving the lowest safety costs, underscoring its superior safety performance.

Additionally, all algorithms were subjected to 50 test iterations in each scenario, with the outcomes detailed in Table \ref{tab:result_compare}. The results validate the effectiveness of our decision-making GPT model, particularly in Scenario 3, which features the most complex traffic conditions. Our method not only reported the lowest instances of safety violations but also the highest rewards, confirming its robustness and superior performance in challenging environments. Meanwhile, the animated version of testing cases are provided, which can be accessed at the site.\footnote{See \url{https://drive.google.com/drive/folders/109kdT4oRNTNztDrw9HzHcb4aBywfTYep?usp=sharing}}

\begin{table}[h]
\centering
\caption{The average evaluation results of different methods in different scenarios.}
\label{tab:result_compare}
% \small
\footnotesize  
\setlength{\tabcolsep}{5pt}
\begin{tabular}{@{}cccccc@{}}
\toprule
Scenario & Metric & BC & BCQL & GPT (w/o ER) & Our Method \\
\midrule
\multirow{2}{*}{Scenario 1} & Reward & 560.64 & 98.42 & 477.27 & \textbf{705.81} \\
                            & Cost   & 41.61 & 10.92 & 22.54 & \textbf{8.06} \\
\multirow{2}{*}{Scenario 2} & Reward & 217.20 & 393.70 & 371.52 & \textbf{662.51} \\
                            & Cost   & 23.19 & 16.60 & 44.78 & \textbf{15.94} \\
\multirow{2}{*}{Scenario 3} & Reward & 527.92 & 456.13 & 327.15 & \textbf{765.73} \\
                            & Cost   & 30.42 & 37.60 & 32.23 & \textbf{5.79} \\
\bottomrule
\end{tabular}
\end{table}

\section{Conclusion}
Decision-making processes are crucial for ensuring the operational integrity and safety of AVs. Existing data-driven decision-making algorithms within this sphere show potential for further enhancement. In this study, we have developed an improved decision-making GPT model, framed within the CMDP and approached as a sequence prediction task. We incorporated entropy regularization, a technique that promotes exploration during training, to refine the optimization target of the GPT model. The efficacy of our decision-making GPT model has been assessed across various driving tasks. Comparative analyses with baseline methods indicate that our model achieves superior performance, especially in terms of safety and overall operational effectiveness.

Looking forward, we plan to integrate online fine-tuning methods into the training regimen of our decision-making GPT model to augment its flexibility and generalization capabilities in increasingly complex scenarios. Additionally, we aim to sample and collect high-quality expert data from diverse canonical scenarios to enhance the guidance provided during the GPT model’s learning process. 

\bibliographystyle{IEEEtran}  
\bibliography{reference}

\end{document}